\documentclass[a4paper,11pt]{article}

\usepackage{geometry}
\usepackage{amsfonts}
\usepackage{biblatex}
\bibliography{references}
\usepackage{microtype}
\usepackage{hyperref}
\usepackage{color}
\usepackage[
    labelsep=endash,
    labelfont=bf
]{caption} 
\usepackage{libertinus}

\usepackage{graphicx}
\usepackage{booktabs}
\usepackage[
    group-separator={,},
    separate-uncertainty=true
]{siunitx} 
\usepackage{multirow}
\usepackage{bigdelim}
\usepackage{bm}
\usepackage[capitalize]{cleveref}

\title{A churn prediction dataset from the telecom sector:\\a new benchmark for uplift modeling}
\author{
  Théo Verhelst\thanks{
  Machine Learning Group, Department of Computer Science, Université Libre de Bruxelles, Brussels, Belgium} \thanks{Corresponding author, \texttt{theo.verhelst@ulb.be}}
  \and
  Denis Mercier\thanks{
  Data Science Team, Orange Belgium, Brussels, Belgium}
  \and
  Jeevan Shrestha\footnotemark[3]
  \and
  Gianluca Bontempi\footnotemark[1]
}
\date{}

\begin{document}
\maketitle

\begin{abstract}
\noindent Uplift modeling, also known as individual treatment effect (ITE) estimation, is an important approach for data-driven decision making that aims to identify the causal impact of an intervention on individuals. This paper introduces a new benchmark dataset for uplift modeling focused on churn prediction, coming from a telecom company in Belgium, Orange Belgium. Churn, in this context, refers to customers terminating their subscription to the telecom service. This is the first publicly available dataset offering the possibility to evaluate the efficiency of uplift modeling on the churn prediction problem. Moreover, its unique characteristics make it more challenging than the few other public uplift datasets.
\end{abstract}

\section{Introduction}
\label{sec:intro}
Uplift modeling, often called the conditional average treatment effect, has become a crucial tool for data-driven decision making. This modeling technique estimates the effect that a particular intervention or treatment has on individuals, enabling the selection of only those individuals who are likely to have a positive reaction to the action. Although the methodology of uplift modeling has witnessed substantial development and diversification~\parencite{gutierrez2016causal}, a notable constraint remains: the low number of publicly available datasets designed specifically for uplift modeling. A recent uplift benchmark conducted by~\textcite{rossler2022bridging} listed only 4 public uplift datasets: Criteo~\parencite{diemerteustache2018large}, Hillstrom~\parencite{hillstrom2008minethatdata}, Starbucks\footnote{\url{https://github.com/joshxinjie/Data_Scientist_Nanodegree/tree/master/starbucks_portfolio_exercise}} and Lenta\footnote{\url{https://www.uplift-modeling.com/en/latest/api/datasets/fetch_lenta.html}}. Furthermore, despite the fact that customer churn is often cited as a common application for uplift modeling, none of these public datasets are concerned with churn. To address this issue, this paper introduces a new churn dataset for uplift modeling, coming from a major telecom company in Belgium, Orange Belgium. This dataset offers researchers and practitioners a new resource to evaluate strategies aimed at reducing churn and increasing customer retention within the telecommunications industry. In \cref{sec:campaigns} we present the marketing campaigns that form the basis of this dataset, and we compare the characteristics of our dataset to two other public datasets in \cref{sec:desc,sec:checks}. Then, in \cref{sec:exp,sec:res} we evaluate the performance of three models on these datasets, and finally we conclude by highlighting its potential to foster innovation and progress within the uplift modeling domain in \cref{sec:conclusion}. The dataset is available on the OpenML platform\footnote{\url{https://www.openml.org/search?type=data&id=45580}} and the benchmark code is available on GitHub\footnote{\url{https://github.com/TheoVerhelst/Churn-Uplift-Dataset-Paper}}.

\section{Churn campaigns}
\label{sec:campaigns}
\begin{figure}
    \centering
    \includegraphics[width=0.8\linewidth]{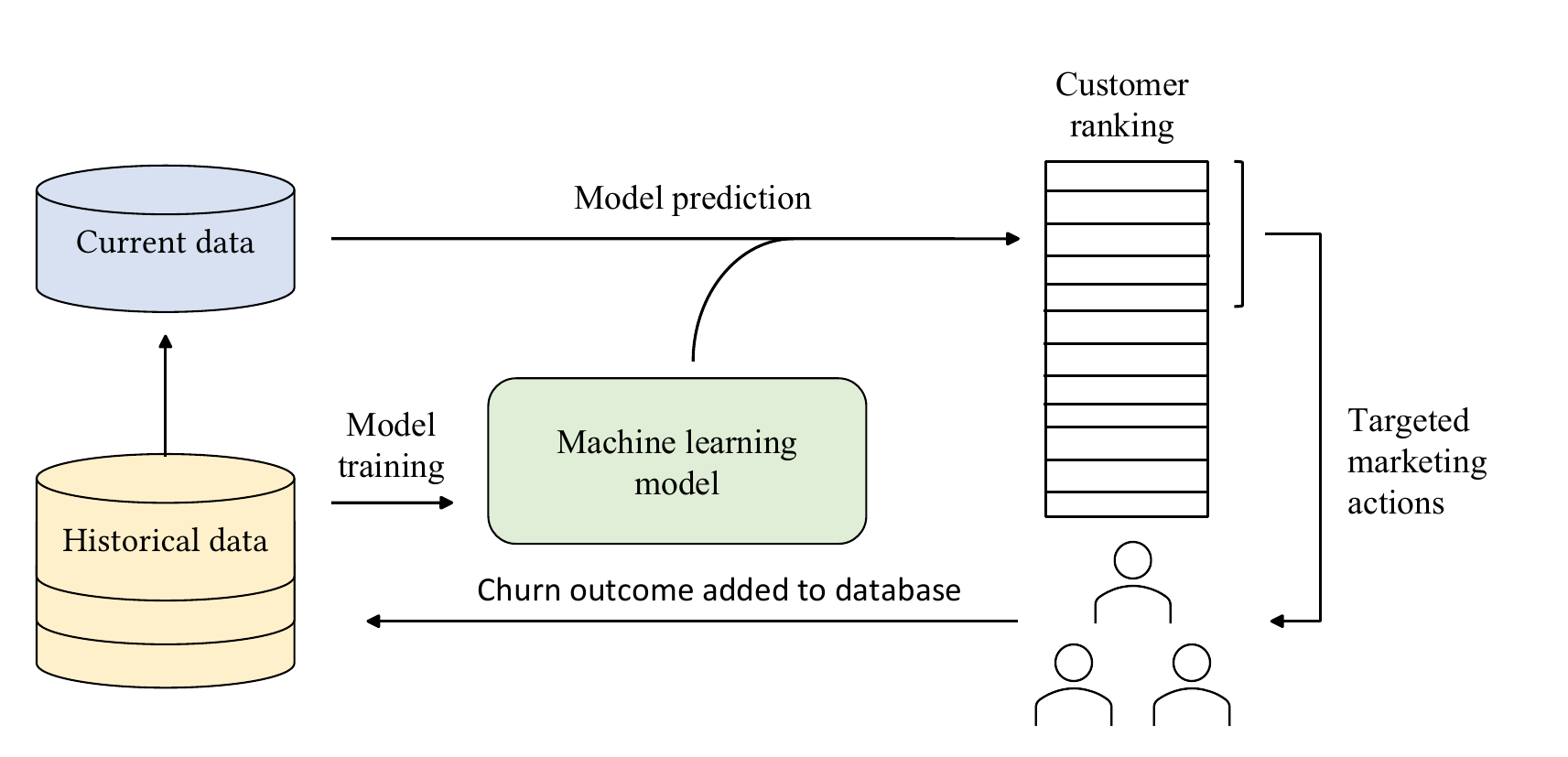}
    \caption{Schematic representation of the churn retention campaign pipeline.}
    \label{fig:pipeline}
\end{figure}

The dataset comes from a series of three marketing campaigns conducted between September and December 2020. The campaign pipeline is represented in \cref{fig:pipeline}. During each campaign, the probability of churn for each customer was estimated using a predictive model and the riskiest customers were selected. A subset of these high-risk customers was randomly assigned to the control group, while the remaining customers formed the target group. The list of customers in the target group was shared with a call center tasked with contacting each customer and presenting them with a marketing offer or recommending a new tariff plan based on their individual history. The churn outcome is determined in a two-month window following the campaign, and any subsequent churn is not attributed to this specific campaign. The data of this campaign and the churn outcome are then recorded in the historical database, and the same campaign process is repeated the next month.

\section{Description}
\label{sec:desc}

\begin{table}
    \centering
    \caption{Description of the churn dataset and two other uplift datasets.}
    \label{tab:datasets}
    \begin{tabular}{lS[table-format=3]S[table-format=8]S[table-format=2.2\%]S[table-format=2.2\%]S[table-format=2.2\%]S[table-format=2.2\%]}
    \toprule
    {Name} & {Features} & {Samples} & \parbox[r]{2.1cm}{\centering Control response rate (\%)} & \parbox[r]{2.1cm}{\centering Target response rate (\%)} & \parbox[r]{2.1cm}{\centering Treatment rate (\%)} \\
    \midrule
    Churn     & 178 & 11896 & 3.6  & 3.4  & 75.74 \\
    Hillstrom & 15  & 42693 & 10.62 & 15.14 & 66.71  \\
    Criteo    & 12  & 25309483 & 4.2   & 4.9   & 84.6 \\
    \bottomrule
    \end{tabular}
\end{table}
The characteristics of the dataset are summarized in \cref{tab:datasets}, as well as the same characteristics of two popular uplift datasets, the Criteo dataset~\parencite{diemerteustache2018large}, and the Hillstrom dataset~\parencite{hillstrom2008minethatdata}. We report the number of features, the number of samples, the response rate in the control and target groups, and the treatment rate.

The churn dataset consists of 11,896 samples, a relatively small number compared to other publicly available uplift datasets. However, it has a larger number of features, totaling 178. These features encompass a diverse range of customer attributes, including demographics (e.g., region of residence, age), usage patterns (e.g., data consumption, number of calls), and subscription details (e.g., price of the tariff plan). The dataset comprises features of various types, including discrete numerical, continuous, and categorical variables, each exhibiting diverse distributions. To ensure privacy and data confidentiality, the dataset is anonymized by using a Principal Component Analysis (PCA) projection of the numerical features, allowing for effective analysis and modeling while protecting sensitive information. Adopting this strategy has proven effective in preserving predictive accuracy while safeguarding privacy in the domain of fraud detection~\parencite{dalpozzolo2015calibrating}. All categorical features and their levels are anonymized by giving them generic names.

One distinctive aspect of the dataset is the inherent difficulty in accurately predicting the churn outcome. The complex dynamics of churn in the telecom sector make it a challenging task, requiring advanced modeling techniques to capture the underlying patterns and factors influencing customer behavior. Uplift modeling is even more difficult than predicting the outcome probability alone due to the relatively small effect of the treatment. To quantify this aspect, we estimate the mutual information $I(\bm x;\bm y_t)$ (for $t=0,1$), which represents the difficulty in predicting the binary outcome $\bm y_t$ from the set of features $\bm x$~\parencite{cover1991elements}. It is estimated using the formula
\[I(\bm x;\bm y_t)=H(\bm y_t)-H(\bm y_t\mid\bm x)\approx H(\bm y_t)-\frac 1N\sum_{i=1}^NH\left(\bm y_t\mid\bm x=x^{(i)}\right)\]
where the term $H(\bm y_t)$ is estimated from the prior distribution of $\bm y_t$, and a T-learner uplift model (details of the experimental setup are presented in \cref{sec:exp}) provides the necessary probability estimates to compute
\[H(\bm y_t\mid x)=P(\bm y_t=0\mid x)\log P(\bm y_t=0\mid x)+P(\bm y_t=1\mid x)\log P(\bm y_t=1\mid x).\]
\begin{table}
    \centering
    \caption{Estimates of the mutual information between the features and the outcomes.}
    \label{tab:mi}
    \begin{tabular}{lS[table-format=.2]S[table-format=.2]S[table-format=.4]S[table-format=.4]S[table-format=2.2\%]S[table-format=2.2\%]}
    \toprule
     & {$H(\bm y_0)$} & {$H(\bm y_1)$} & ${\hat I(\bm x; \bm y_0)}$ & ${\hat I(\bm x; \bm y_1)}$ & ${\frac{\hat I(\bm x; \bm y_{0})}{H(\bm y_0)}}$ & ${\frac{\hat I(\bm x; \bm y_{1})}{H(\bm y_1)}}$ \\
    \midrule
    Churn     & .16 & .15 & .0008 & .0025 & 0.54\%  & 1.71\%  \\
    Hillstrom & .34 & .43 & .0112 & .0123 & 3.32\%  & 2.90\%  \\
    Criteo    & .16 & .20 & .0429 & .0573 & 24.63\% & 29.32\% \\
    \bottomrule
    \end{tabular}
\end{table}
The estimates of the mutual information are given in \cref{tab:mi}. In the last two columns, the mutual information estimate is divided by the entropy of the prior distribution, indicating the proportion of uncertainty of the outcome explained by the features. Our dataset has a low outcome probability similar to that of the Criteo dataset, while also having a very low mutual information, like the Hillstrom dataset. This represents a unique contribution to the uplift ecosystem, where there are no other datasets that come from a small-scale marketing campaign with these particular characteristics.

To better characterize the differences in the outcome distribution among the three datasets, we use the counterfactual point estimator proposed in~\parencite{verhelst2023partial}. It estimates the probabilities of the joint distribution of the potential outcomes $\bm y_0,\bm y_1$, even though this distribution cannot be observed directly. We use the following formula, also based on the probability estimates given by the T-learner:
\begin{equation}
    P(\bm y_0=y_0,\bm y_1=y_1)\approx\frac 1N\sum_{i=1}^NP\left(\bm y_0=y_0\mid\bm x^{(i)}\right)P\left(\bm y_1=y_1\mid\bm x^{(i)}\right).
\end{equation}
for $y_0,y_1\in\{0,1\}$.
\begin{table}
    \centering
    \caption{Estimated distribution of counterfactuals.}
    \label{tab:cf}
    \begin{tabular}{lll*{3}{S[table-format=2.1\%]}}
    \toprule
    Formula & {Name (in churn)}  & {Name (in retail)} &{Churn} & {Hillstrom} & {Criteo} \\
    \midrule
    ${P(\bm y_0=0,\bm y_1=0)}$ & Sure thing & Lost cause & 93.1\% & 76.0\% & 90.8\% \\
    ${P(\bm y_0=1,\bm y_1=0)}$ & Persuadable & Do-not-disturb & 3.5\% & 8.9\% & 2.8\% \\
    ${P(\bm y_0=0,\bm y_1=1)}$ & Do-not-disturb & Persuadable & 3.2\% & 13.4\% & 4.8\% \\
    ${P(\bm y_0=1,\bm y_1=1)}$ & Lost cause & Sure thing & 0.1\% & 1.8\% & 1.7\% \\
    \bottomrule
    \end{tabular}
\end{table}
Estimated probabilities are reported in \cref{tab:cf}, along with the business name associated with the four counterfactuals. Note that, in the churn setting, the outcome $\bm y=1$ should be avoided, while in retail, its probability should be maximized. This implies that the two contexts associate different names to the same probabilities. We see that the churn dataset is characterized by a low probability that both potential outcomes are $1$ (\emph{lost cause} customers, fourth row). This suggests that a negligible number of customers are likely to churn regardless of the targeted marketing action. This differs from the two other datasets, for which this probability is higher. We also observe that the churn dataset is more balanced between positive and negative causal effects (second and third rows), whereas, in both the Hillstrom and Criteo datasets, there is a larger proportion of individuals with a positive causal effect (\emph{persuadable} customers, third row).

\section{Randomization}
\label{sec:checks}
\begin{figure}
    \centering
    \includegraphics[width=8cm]{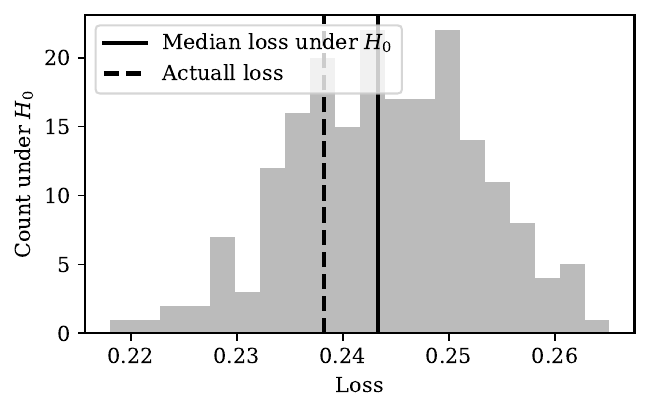}
    \caption{Distribution of the loss under the hypothesis that the treatment is randomized.}
    \label{fig:treatment_test}
\end{figure}

Since the dataset comes from a randomized campaign, the treatment should be independent of the outcomes. To validate this independence, we performed the Classifier 2 Sample Test used in~\parencite{diemerteustache2018large}. We trained a classifier to predict the treatment indicator and compared its Hamming loss with the loss distribution obtained under the null hypothesis, sampled by training models to predict random splits. The treatment predictor has a loss of 23.82\% (close to the proportion of control samples, 24.26\%), which corresponds to a p-value of 0.26 under the null hypothesis. This result is shown in \cref{fig:treatment_test}. This indicates that the treatment cannot be predicted based on available features, hence the randomization of treatment assignment can be considered appropriate and unbiased.

\section{Benchmark experimental setup}
\label{sec:exp}
We conducted an experimental benchmark on the churn, Hillstrom and Criteo datasets. We used the classical random forest (RF) model~\parencite{breiman2001random}, the T-learner uplift model~\parencite{kunzel2019metalearners}, and the uplift random forest~\parencite{guelman2015uplift}. The classical RF model, which we call outcome RF, was trained to predict churn with control samples, therefore, without explicitly considering the individual treatment effect. This serves as a baseline for evaluating the performance of uplift models. This is especially relevant in sectors such as telecoms, where such predictive models are often used instead of uplift models because of their simplicity and sufficient performance. The T-learner used a random forest as base learner. All three models used 100 trees, a maximum depth of 20, and a minimum of 10 samples per leaf. To tackle class imbalance, the EasyEnsemble strategy~\parencite{liu2009exploratory} was applied with 8 folds. For each fold, a new model was trained using all positive samples and an equally sized set of randomly selected negative samples. This approach helps mitigate the adverse impact of class imbalance. The predictions of the eight models were then averaged, effectively reducing the potential biases caused by undersampling the negative samples in individual folds. K-fold cross-validation with $k=3$ was used to obtain training and test splits of each dataset. Finally, the whole experiment was repeated 10 times to obtain a more robust estimation of the performance, as well as an estimation of its variability. The performance of each model was estimated in terms of the area under the uplift curve (AUUC)~\parencite{gubela2021uplift}.

\section{Results}
\label{sec:res}
The AUUC is reported in \cref{tab:auuc}. To evaluate the impact of the PCA projection, we performed the same experiment on the original, non-anonymized churn dataset. It appears that the performance is only slightly lower on the anonymized dataset than on the original, and, given the high uncertainty in the AUUC, we cannot exclude that this difference is due to random variations in the benchmark sampling. We also observe that the performance of all models is highest on the Hillstrom dataset and lowest on the churn dataset. We attribute this difference to the fact that the outcome is balanced in the Hillstrom dataset, whereas it is unbalanced and more difficult to predict in the churn dataset. Interestingly, the performance of the outcome RF model is consistently the highest, showing that the uplift approach is not always preferable, as discussed in~\parencite{fernandez-loria2022causal,fernandez-loria2022causala}.

\begin{table}
    \centering
    \caption{Mean and standard deviation of the area under the uplift curve (AUUC) in the benchmark.}
    \label{tab:auuc}
    \begin{tabular}{l*{4}{S[table-format=1.2(1.1)]}}
        \toprule
        & {Churn (\%)} & \parbox[r]{2.7cm}{\centering Churn\\(not anonymized) (\%)} & {Hillstrom (\%)} & {Criteo (\%)} \\
        \midrule
        Outcome RF   & 0.26(0.47) & 0.33(0.37) & 2.20(0.32) & 1.01(0.25) \\
        Uplift RF    & 0.19(0.37) & 0.22(0.29) & 2.19(0.27) & 0.89(0.23) \\
        T-learner RF \hspace{1em} & 0.25(0.38) & 0.33(0.39) & 2.72(0.28) & 0.86(0.18) \\
        \bottomrule
    \end{tabular}
\end{table} 

The estimator variance likely plays an important role in determining when classical predictive modeling outperforms uplift modeling~\parencite{fernandez-loria2022causal,fernandez-loria2022causala}. To evaluate this possibility, we computed the variance of the predicted probability estimates of each model on each data sample. This was achieved by considering the 10 different predictions generated for each sample during the repeated 3-fold cross-validation procedure. The values reported in \cref{tab:variance} represent the variance averaged across all samples in the dataset. We observe that the two uplift models in this benchmark suffer from a higher variance than the outcome RF, especially on the Criteo dataset.

\begin{table}
    \centering
    \caption{Variance of the predictions averaged over the dataset.}
    \label{tab:variance}
    \begin{tabular}{l!{\;}*{3}{S[table-format=1.2e-1]!{\;}}}
    \toprule
    & {Churn} & {Hillstrom} & {Criteo} \\
    \midrule
    Outcome RF   & 2.07e-3 & 3.49e-3 & 1.22e-3  \\
    Uplift RF    & 3.06e-3 & 4.33e-4 & 2.05e-3  \\
    T-learner RF & 3.78e-3 & 7.59e-3 & 1.94e-3 \\
    \bottomrule
    \end{tabular}
\end{table}

\section{Conclusion}
\label{sec:conclusion}
The primary objective of this new dataset is to facilitate the evaluation and comparison of uplift modeling techniques, with a focus on customer churn prediction in the telecom sector. More generally, researchers and practitioners can leverage this dataset to develop and benchmark new algorithms, feature engineering approaches, and model evaluation metrics tailored to uplift modeling in difficult settings characterized by a low information rate, a low outcome probability, and a small number of samples. This is especially interesting for smaller companies aiming to initiate personalized marketing campaigns, but which have limited historical data to train uplift models. While large-scale benchmarks such as the Criteo dataset are crucial for evaluating the performance of uplift models on a large sample, small-scale datasets are more representative of some practical applications. This dataset also provides an opportunity to assess other causal inference methods such as counterfactual estimation~\parencite{li2019unit,verhelst2023partial}. Finally, we observed in a benchmark experiment that classical predictive modeling is more effective than uplift modeling~\parencite{fernandez-loria2022causal,fernandez-loria2022causala}. This has also been observed in practice by our industrial partner. In future work, we intend to investigate this question from a theoretical perspective with the hope of gaining a deeper understanding of the critical factors that impact the performance of both predictive and uplift approaches.

\printbibliography

@article{rossler2022bridging,
	title = {Bridging the gap: {A} systematic benchmarking of uplift modeling and heterogeneous treatment effects methods},
	volume = {57},
	number = {4},
	journal = {Journal of Interactive Marketing},
	author = {Rößler, Jannik and Schoder, Detlef},
	year = {2022},
	note = {Publisher: SAGE Publications Sage CA: Los Angeles, CA},
	pages = {629--650},
}

@article{hillstrom2008minethatdata,
	title = {The minethatdata e-mail analytics and data mining challenge, 2008},
	journal = {URL https://blog. minethatdata. com/2008/03/minethatdata-e-mail-analytics-and-data. html},
	author = {Hillstrom, Kevin},
	year = {2008},
}

@article{verhelst2023partial,
	title = {Partial counterfactual identification and uplift modeling: theoretical results and real-world assessment},
	copyright = {All rights reserved},
	issn = {0885-6125, 1573-0565},
	shorttitle = {Partial counterfactual identification and uplift modeling},
	url = {https://link.springer.com/10.1007/s10994-023-06317-w},
	doi = {10.1007/s10994-023-06317-w},
	language = {en},
	urldate = {2023-05-03},
	journal = {Machine Learning},
	author = {Verhelst, Théo and Mercier, Denis and Shrestha, Jeevan and Bontempi, Gianluca},
	month = mar,
	year = {2023},
}

@article{gubela2021uplift,
	title = {Uplift modeling with value-driven evaluation metrics},
	journal = {Decision Support Systems},
	author = {Gubela, Robin M and Lessmann, Stefan},
	year = {2021},
	note = {Publisher: Elsevier},
	pages = {113648},
}

@inproceedings{li2019unit,
	title = {Unit {Selection} {Based} on {Counterfactual} {Logic}},
	url = {https://doi.org/10.24963/ijcai.2019/248},
	doi = {10.24963/ijcai.2019/248},
	booktitle = {{IJCAI}},
	publisher = {International Joint Conferences on Artificial Intelligence Organization},
	author = {Li, Ang and Pearl, Judea},
	year = {2019},
	pages = {1793--1799},
}

@article{guelman2015uplift,
	title = {Uplift random forests},
	volume = {46},
	issn = {10876553},
	doi = {10.1080/01969722.2015.1012892},
	number = {3-4},
	journal = {Cybernetics and Systems},
	author = {Guelman, Leo and Guillén, Montserrat and Pérez-Marín, Ana M.},
	year = {2015},
	note = {Publisher: Taylor \& Francis},
	pages = {230--248},
}

@article{breiman2001random,
	title = {Random forests},
	volume = {45},
	number = {1},
	journal = {Machine learning},
	author = {Breiman, Leo},
	year = {2001},
	note = {Publisher: Springer},
	pages = {5--32},
}

@article{kunzel2019metalearners,
	title = {Metalearners for estimating heterogeneous treatment effects using machine learning},
	volume = {116},
	issn = {10916490},
	doi = {10.1073/pnas.1804597116},
	number = {10},
	journal = {Proceedings of the National Academy of Sciences of the United States of America},
	author = {Künzel, Sören R. and Sekhon, Jasjeet S. and Bickel, Peter J. and Yu, Bin},
	year = {2019},
	pmid = {30770453},
	note = {arXiv: 1706.03461
Publisher: National Acad Sciences},
	pages = {4156--4165},
}

@article{liu2009exploratory,
	title = {Exploratory undersampling for class-imbalance learning},
	volume = {39},
	doi = {10.1109/tsmcb.2008.2007853},
	number = {2},
	journal = {IEEE Transactions on Systems, Man, and Cybernetics, Part B (Cybernetics)},
	author = {Liu, Xu-Ying and Wu, Jianxin and Zhou, Zhi-Hua},
	year = {2009},
	note = {Publisher: IEEE},
	pages = {539--550},
}

@book{cover1991elements,
	title = {Elements of information theory},
	isbn = {0-471-06259-6},
	publisher = {John Wiley \& Sons},
	author = {Cover, Thomas M. and Thomas, Joy A.},
	year = {1991},
	doi = {10.1002/0471200611},
	note = {Publication Title: Elements of Information Theory},
}

@article{fernandez-loria2022causal,
	title = {Causal {Classification}: {Treatment} {Effect} {Estimation} vs. {Outcome} {Prediction}},
	volume = {23},
	number = {59},
	journal = {Journal of Machine Learning Research},
	author = {Fernández-Loria, Carlos and Provost, Foster},
	year = {2022},
	pages = {1--35},
}

@article{fernandez-loria2022causala,
	title = {Causal decision making and causal effect estimation are not the same… and why it matters},
	journal = {INFORMS Journal on Data Science},
	author = {Fernández-Loria, Carlos and Provost, Foster},
	year = {2022},
	note = {Publisher: INFORMS},
}

@inproceedings{gutierrez2016causal,
	address = {Microsoft NERD, Boston, USA},
	title = {Causal {Inference} and {Uplift} {Modelling}: {A} {Review} of the {Literature}},
	volume = {67},
	url = {http://proceedings.mlr.press/v67/gutierrez17a.html},
	booktitle = {Proceedings of {The} 3rd {International} {Conference} on {Predictive} {Applications} and {APIs}},
	publisher = {PMLR},
	author = {Gutierrez, Pierre and Gérardy, Jean-Yves},
	editor = {Hardgrove, Claire and Dorard, Louis and Thompson, Keiran and Douetteau, Florian},
	month = jan,
	year = {2016},
	note = {Series Title: Proceedings of Machine Learning Research},
	pages = {1--13},
}

@inproceedings{dalpozzolo2015calibrating,
	title = {Calibrating probability with undersampling for unbalanced classification},
	booktitle = {2015 {IEEE} {Symposium} {Series} on {Computational} {Intelligence}},
	publisher = {IEEE},
	author = {Dal Pozzolo, Andrea and Caelen, Olivier and Johnson, Reid A and Bontempi, Gianluca},
	year = {2015},
	pages = {159--166},
}

@inproceedings{diemerteustache2018large,
	title = {A {Large} {Scale} {Benchmark} for {Uplift} {Modeling}},
	booktitle = {Proceedings of the {AdKDD} and {TargetAd} {Workshop}, {KDD}, {London},{United} {Kingdom}, {August}, 20, 2018},
	publisher = {ACM},
	author = {Diemert Eustache, Betlei Artem and Renaudin, Christophe and Massih-Reza, Amini},
	year = {2018},
}

\end{document}